\documentclass{article}
\usepackage{graphicx} 

\usepackage{graphicx}
\usepackage[T1]{fontenc}
\usepackage{xcolor}
\usepackage{float}
\usepackage{amsmath}
\usepackage{comment}
\usepackage{url}
\usepackage[section]{placeins}
\usepackage{algorithm}
\usepackage{algpseudocode}
\usepackage{multirow}
\usepackage{tikz}
\usetikzlibrary{calc}
\usepackage{subcaption}

\title{Data Classification with Dynamically Growing and Shrinking Neural Networks\thanks{
A. Jastrzebska's contribution was supported by the National Science Centre, grant No. 2024/53/B/ST6/00021.}}

\author{Szymon \'Swiderski$^a$ \and
Agnieszka Jastrz\k{e}bska$^{a,b}$
}
\date{$^a$Warsaw University of Technology, Warsaw, Poland\\$^b$The John Paul II Catholic University of Lublin, Poland}

\begin{document}
\maketitle

\begin{center}
    \textbf{Abstract}\\
\end{center}

The issue of data-driven neural network model construction is one of the core problems in the domain of Artificial Intelligence. A~standard approach assumes a fixed architecture with trainable weights. A~conceptually more advanced assumption is that we not only train the weights, but also find out the optimal model architecture. We present a~new method that realizes just that.  This article is an extended version of our  conference paper titled ``Dynamic Growing and Shrinking of Neural Networks with Monte Carlo Tree Search \cite{ICCS24_ORGINAL}''.
In the paper, we show in detail how to create a neural network with a procedure that allows dynamic shrinking and growing of the model while it is being trained. The decision-making mechanism for the architectural design is governed by a Monte Carlo tree search procedure which simulates network behavior and allows to compare several candidate architecture changes to choose the best one. 
The proposed method was validated using both visual and time series datasets, demonstrating its particular effectiveness in multivariate time series classification. This is attributed to the architecture's ability to adapt dynamically, allowing independent modifications for each time series. The approach is supplemented by Python source code for reproducibility.  Experimental evaluations in visual pattern and multivariate time series classification tasks revealed highly promising performance, underscoring the method's robustness and adaptability.
\newpage

\section{Introduction}

Neural network training algorithms development is an essential theoretical and practical problem of Artificial Intelligence. The underlying task is network architecture design. Typical methods assume that a programmer specifies subsequent components of the network and uses an optimization algorithm of choice to find out weight values. At the same time, there is a pressing need to deliver effective methods that relieve programmers from the task of neural network architecture specification. One trend is to use predefined designs, tested by others on some benchmark datasets. A more conceptually advanced scenario is to offer training algorithms that optimize the network architecture during the training procedure. In this scenario, the training algorithm is responsible not only for setting the weights but also for modifying the model design.

The idea of delegating neural model design to an optimization algorithm has been present in the literature domain for some time now. Unfortunately, up to this day, the practical use of this approach is quite limited. This is first and foremost due to the modest effectiveness of the available approaches. Most of the existing studies, such as the ones of Zhang et al. \cite{Zhang2022}, were delivered for plain feed-forward neural networks. The demands of contemporary data analysis, especially in the field of image classification, are not matched when such models are used. There are some studies, such as the very recent paper by Evci et al. \cite{GrandMax}, that offer more insights. The aforementioned paper shows an approach when a network is grown/shrunk neuron-by-neuron. Comprehensive theoretical and empirical studies are overall rare. 

In light of the facts outlined in the previous paragraph, in this paper, we contribute a novel method for neural network training. We are publishing it as an open-source package. We chose the name $growingnn$. It was uploaded to PyPi. Its description is under \url{https://pypi.org/project/growingnn/}. The delivered method uses error backpropagation as a base for weight update. The action of architecture design is carried out by enabling a scheduler that after each $K$~epochs allows the neural architecture to change. The change can be realized by adding or removing a layer of a predefined type from the network. The current implementation covers three types of neural layers: (i)~a~plain, feed-forward layer, (ii)~a~convolutional layer, and (iii)~a~layer with residual connections. These three layer types are typical in contemporary advanced models for visual pattern recognition.

This article is an extension of our published conference paper titled ``Dynamic Growing and Shrinking of Neural Networks with Monte Carlo Tree Search \cite{ICCS24_ORGINAL}''. The original study has been expanded to include data related to time series analysis. The method presented here shows great potential with time series data because each series can evolve independently. During training, the structure of the network dedicated to analyzing each individual time series adjusts dynamically, leading to a more optimized and efficient solution. This feature makes the method especially suited for handling complex time series problems.

The key novelty of the presented work is the use of Monte Carlo tree search to simulate network performance. We use it to determine the optimal decision with regard to the design change. To achieve that, the neural model evolution strategy is represented as a tree. The network is redesigned in such a way that a change in the structure has a minimal impact on the already-learned weight values. In this manner, instead of offering a trivial wrapper solution that trains from scratch a new model using a library algorithm after an architecture update, we develop a novel optimizer that performs continual (progressive) training and reuse of the already-learned neural connections. Furthermore, the changes to the network architecture concern entire layers, not single neurons.

The proposed approach was empirically evaluated in applied visual pattern recognition tasks using standard benchmark datasets: MNIST \cite{deng2012mnist} and FMNIST \cite{FMNIST}. Additionally, the method was tested on time series data from the benchmark introduced in the  study titled ``The Great Multivariate Time Series Classification Bake Off'' by Ruiz et al. \cite{Ruiz2021}. The performance of the model was compared with baseline strategies for neural architecture modification, including random and greedy approaches. In all cases, including both visual and time series data, the Monte Carlo tree search method achieved significantly better results than the alternative strategies.

Our proposed approach, in its current state, is suitable for convolutional neural networks. In this setting, created models are ideal for image classification. The input data in this problem are available in the form of two (or more) dimensional matrices, in which values encode pixel colors. We want to emphasize that operating on such input does not restrict us from exclusively processing images. Other data types are often converted to image-like formats. For example, sound waves can be converted to spectrograms. Also, time series can be converted to image-based representations.

The task of classifying multivariate time series is a complex challenge that necessitates sophisticated models and advanced machine learning methods \cite{MCTS_general}. A multivariate time series dataset comprises multiple variables, each represented by its own time series, with the potential for varying lengths and sampling rates across these series \cite{JOCS_MCTS,JOCS_MCTS2}.
In this paper, we show the application of the proposed model to both image and time series data processing.

The remainder of this paper is structured as follows. Section \ref{sec:literature} addresses relevant literature positions in the domain of architecture-changing neural network training methods. Section \ref{sec:theory} outlines the theoretical background of our approach. Section \ref{sec:empiricalresults} shows the results of empirical tests of the new method. Section \ref{sec:mtsc} shows the application of the developed model to multivariate time series classification. Section \ref{sec:conclusion} concludes the paper.

\section{Brief literature survey}
\label{sec:literature}
In recent years, the topic of dynamic neural network architecture change has attracted noticeable attention. We shall start this discussion by mentioning the method known as GradMax \cite{GrandMax}. It is a method capable of growing a neural architecture during the training procedure without costly retraining. The idea is very similar to the one discussed in this paper, but the methods that handle each change are very different from our methods. GradMax operates on the level of a single neuron. In our algorithm, there is a very wide spectrum of possible changes for a network that allows architecture to grow and shrink. GradMax maximizes gradients for new weights and efficiently initializes them using singular value decomposition (SVD). This approach makes new neurons not impact existing knowledge which is contradictory to our method for which neural network has a short period of instability. The idea of changing the structure using gradient information is relatively common in the literature. One of the first of this kind was a model called resource-allocating network  \cite{Plat_GNN}. In this method, when a given pattern was unrecognizable, new neurons were added. An analogous idea was published by Fahlman and Lebiere under the name Cascade-Correlation Architecture   \cite{Cascore}. Neural networks can grow not only by adding new neurons but also by splitting existing neurons. The article by Kilcher et al. \cite{kilcher2019escaping} about escaping flat areas via structure modification introduces a new strategy of this kind. When the change of loss is slowing down and the network encourages a flat error surface, the proposed method adds new neurons by splitting the existing ones. This work has two important elements in common with our method. The changes to the structure are made when the network’s ability to learn decreases and we also believe that their method of splitting the neuron is in a few ways similar to our method that uses quasi-identity matrices. A very important question in the field of neural networks that can change their structure is how much the neural network can adapt to the problem. A study about Convex Neural Networks \cite{Bach14} shows that these networks can adapt to diverse linear structures by adding neurons in a single hidden layer in each step of training, which also forces the network to grow.

Modern machine learning problems, such as Transformer-based models, often benefit from integrating convolutional layers, especially in the case of Vision Transformers  \cite{JOCS_WANG2023102121}. Creating such complex architectures is a big challenge because Vision Transformers combine convolutional and self-attention mechanisms to process local and global information efficiently. This balance of local and global feature processing inspired us to develop a tool designed to optimize structures with convolutional layers.

\section{The method}
\label{sec:theory}
The proposed algorithm consists of two components. The first component performs weight adjustment. 
The second component is the orchestrator, which launches a procedure to change the network architecture. The change takes place each $K$ epoch and it works in a guided manner. Its decisions are made based on the outcome of the Monte Carlo tree search. A rough outline of the new routine is given in the Algorithm \ref{alg:general_training}.

The proposed algorithm consists of two components. The first component performs weight adjustment. The second component is the orchestrator, which launches a procedure to change the network architecture. The change takes place each K epoch and it works in a guided manner. Its decisions are made based on the outcome of the Monte Carlo tree search. A rough outline of the new routine is given in the Algorithm \ref{alg:general_training}.

The algorithm is fundamentally built upon Stochastic Gradient Descent (SGD) \cite{StochasticGradientDescent} \cite{JOCS_OTIMIZATION}. It relies on low-level computations, avoiding elaborate tools used in some contemporary training algorithms. This simplicity makes it ideal for research that focuses on fundamental machine-learning principles. For us, a model is a main structure that stores layers as nodes in a directed graph structure, it operates on the identifiers of layers. The layer is assumed to be an independent structure that contains information about incoming and outgoing connections. The default starting structure is a graph that has an input and output layer and one connection between those. In each generation, a new layer may be added or an existing layer may be removed. As the structure grows, each layer gains more incoming and outgoing connections. In the propagation phase, the layer waits until it receives signals from all input layers. Once received, the signals are averaged, processed, and propagated through all outgoing connections.

\begin{algorithm}[H]
    \caption{Training algorithm.}\label{alg:general_training}
    \begin{algorithmic}[1]
        \State \textbf{Input:} Dataset
        \State \textbf{Output:} Model
        
        \State \textbf{Initialization:}
        \State SimSet $\gets$ Create simulation set
        \State Model $\gets$ Create a model with basic structure
        
        \For{each generation}
            \State GradientDescent(Model, Dataset, epochs)
            \If{canSimulate()}
                \State Action $\gets$ MCTS.simulation(Model)
                \State Model $\gets$ Action.execute(Model)
            \EndIf
        \EndFor
    \end{algorithmic}
\end{algorithm}

The function $canSimulate()$  called in Line 8 in Algorithm \ref{alg:general_training} represents a module that is later referred to as simulation orchestrator. The orchestrator determines the point in the training procedure when a simulation is executed during the learning process to change a current neural architecture. At the end of each generation, the simulation orchestrator checks if a simulation is needed. The moment at which the simulation is executed is very important because it helps maintain a balance between the exploration and exploitation of potential structures. A model may retain a particular architecture for several epochs, or it may require frequent changes. Too frequent changes may prevent a specific architecture from being fully trained, while infrequent changes may lead to constantly running into local minima and significantly increase learning time, rendering the method inefficient. In the section dedicated to parameter exploration, we examined various approaches, but ultimately, we decided to use a method known as progress check. In this method, after each generation, we check whether there has been an improvement in the model’s learning. If there is no improvement, then the simulation is run.

In our algorithm, the learning rate plays a crucial role. We implemented a custom modification of the progressive learning rate in line with the work of Schaul et al. \cite{DBLP:journals/corr/Smith15a}. In this approach, the progressive learning rate ensures that the learning rate is very close to zero in the first epoch after the structure change. Thereafter, the learning rate grows to a constant value through the training process. When the maximum constant value of the learning rate is reached, the learning rate slowly decreases before the next action. This minimizes the negative impact of introduced changes on the already learned information in the network. Neural networks continue to increase in size and complexity, and their energy consumption becomes a significant issue \cite{JOCS_RAJOVIC2013439}. The method proposed in this paper has great potential to reduce the architectural size of neural networks, thereby decreasing their energy requirements without compromising performance. This approach is crucial for making large-scale neural networks more sustainable and practical for real-world applications. There is a growing interest in specialized architectures for specific data types and tasks. Because the general idea for this algorithm is straightforward, we believe this method can be used to optimize even very specialized architectures  \cite{JOCS_PANDEY2024102401}.

\subsection{Neural architecture design changes}

The algorithm draws inspiration from the achievements of ResNet-50 \cite{RESNET50} and ResNeXt \cite{RESNEXT}, particularly from the success of residual connections. The model's structure is treated as a graph in the presented algorithm, where layers are nodes and connections between layers are directed edges. While our model is not a graph neural network, it draws significant inspiration from the principles and methodologies employed in GNNs \cite{JOCS_MAURYA2022101695}. This method does not use a structure built from combining residual and sequential connections to shape a directed acyclic graph, which is similar to spiking neural networks \cite{JOCS_SKIPE}.

The algorithm allows adding and removing layers from the model without losing the residual structure of layers. All layers operate asynchronously, and signals move through the network in a manner akin to recursion. Such a structure has key properties, as it allows for unlimited network size, ensures that data always flows through the network without supervision, prevents deadlocks, and
prohibits cyclic or unnecessary layers.

In general, a network has a tendency to add new layers, which allows the network to grow and learn new features. Changes to the network structure are added in the form of actions. An action concerns essentially either an addition or a removal of a layer. The algorithm generates all of the possible candidate connections for a given type of layer. Each possible connection for a given layer type in structure defines one single action that can be run on the current model. 

The current implementation supports four different types of layers.
\vspace{-5pt}
\begin{enumerate}
	\item sequential dense layer;
	\item residual dense layer;
	\item sequential convolution layer;
	\item residual convolution layer.\vspace{-5pt}
\end{enumerate}

Dense layers can be connected to and from any layer kind as long as a residual structure is preserved. Convolution layers have a specific rule they can only be added after another convolution layer as input. In our experiments, the initial layer is always convolutional as the model is primarily developed to deal with computer vision tasks.

\subsection{Monte Carlo Tree Search}

Monte Carlo Tree Search (MCTS) \cite{montecarlo_overview}  is a search algorithm used in decision processes, particularly in games and simulations. It builds a tree structure by simulating different possible moves, evaluating their outcomes, and expanding the tree. Each iteration in this simulation is divided into four parts.

\begin{description}
    \item[\textbf{1. Selection.}] Starting from the root of the tree, one child is selected. The main  difficulty is to maintain a balance between exploration and exploitation. This     balance is controlled by upper confidence bound applied to trees~\cite{UCB1}:
    \begin{equation}
    a^* = \arg \max_{a \in A(M_s)} \left( Q(M_s, a) + C_s \cdot \sqrt{\frac{\ln N(s)}{N(M_s, a)}} \right).
    \end{equation}
    For a given set of actions $A(s)$ generated for a current model structure $M_s$, the formula selects the action chosen in the child node during the selection. $Q(M_s, a)$ denotes the average result of scores from the rollout phase. $N(s)$ is a number denoting how many times model structure $M_s$ has been analyzed. $N(M_s, a)$ denotes the number how many times action a has been processed     for model structure $M_s$.
   Initially, the root node consists of a model structure    for the current generation.
    \item[\textbf{2. Expansion}.] The algorithm adds new children of a node. It executes all possible actions for a model structure, creating a set of children nodes, each     having a different model structure.
    \item[\textbf{3. Rollout}] or playoff. For a given node which is a leaf, in the simulation tree, the     algorithm is trying to play a random game. In our adaptation, the rollout     executes $n$ random actions on a given model, to simulate future possible    changes after a given action. After the rollout, the resulting structure is    passed to the score function. The score function trains the resulting structure
    on the simulation dataset and the resulting accuracy represents the score     from the rollout.
    \item[\textbf{4. Backpropagation.}] All nodes are updated according to the score function after the rollout.
\end{description}

MCTS in this implementation is time-limited \cite{JOCS_MCTS}. After a user-specified time, the simulation returns a single action. The assumption is that the action performed on the current structure should set the stage for subsequent actions to converge toward an optimal structure. In each generation, MCTS identifies changes in the structure toward an optimal configuration. This behavior is anal-
ogous to the gradient descent algorithm, which in each epoch determines the change in weight for improvement.

For each structure, the algorithm has seven possible action types to generate\vspace{-3pt}
\begin{enumerate}
    \item \textbf{Action: Add a dense sequential layer.} Executing this action adds a sequential layer between two other layers. Unlike residual layers, the sequential     layer does not have the ability to determine the initial state of weights in     the layer.
     \item \textbf{Action: Add a dense residual random layer.}  A residual layer between      two layers does not need to be added between layers that are directly connected. A residual layer can be added between any two layers between which      there is a path in the same direction. The initial state of the weights is very      important because it determines what will happen to the network’s knowledge after adding a new layer.
     \item \textbf{Action: Add a dense residual ``zero'' layer.}  This action involves adding        a residual layer, for which the initial value of all weights is zero. Since the        weights are zero, the residual layer should have almost no effect on the network output in the first forward propagation execution after adding this       layer. The assumption is that the network may be at a point in space where       it cannot move towards the global minimum, as it lacks a dimension in which       it could move.
    \item \textbf{Action: Add a dense residual identity layer.}  The idea in adding this     layer is that the value of the input layer to this layer is enhanced. The weight     matrix in this layer is an identity matrix and the bias values are zero, which     means that the output from this layer is the same as the output from the     input layer to the newly added layer because it is a residual layer.    
    \item \textbf{Action: Add a sequential convolution layer.} Adding a sequential convolution layer works the same as it was in a dense layer, but convolution    layers can be only connected to another convolution layer as input, as output it can be a convolution or dense layer. In the convolution type of actions,    there are no predefined weights initial state.    
    \item \textbf{Action: Add a residual convolution layer.} Adding a residual convolution layer works the same as it was in a dense residual layer, with the same    constraints as it was with convolution sequential action.    
    \item \textbf{Action: Remove a layer.}  Removing a layer cannot change the main principles in the structure, the graph that the layers create must be directed    and acyclic. The algorithm allows to removal of any layer except the initial    input and output layer of the model. The algorithm may create additional    connections to maintain the established structure.
\end{enumerate}
\vspace{-3pt}

Before executing the method, a default number of neurons in layers, denoted as \( def_{neu} \), can be set. This default value does not force all layers to have the same number of neurons, but rather most of them will align with it. Initially, the output layer will have the number of incoming connections set to \( def_{neu} \), which subsequently influences most layers to adopt this neuron count. This alignment occurs because each added layer adapts to the layers it connects to.

\subsection{Training scheme -- the complete algorithm}
In the presented algorithm, iterations for training the model are divided into generations and then into epochs. In every generation, the structure of the model can change. In every epoch, the model changes its weights. For each generation, the algorithm runs a gradient descent algorithm for a specified number of epochs. While training learning rate changes progressively. For the first and last epoch in a single generation, the learning rate is close to zero. In the middle of training, the learning rate gets to some maximal value, this approach makes it easier for the structure to adapt to new changes in the network. Although the key properties of gradient descent are preserved, there is a big change in data flow in forward and backward propagation. If a layer has more than one input signal, it waits until all information is gathered, after that all input signals are averaged and then processed. 

The biggest advantage of the residual structure is that there is no need to supervise the data flow. When a signal is sent to the input layer of the model, it is guaranteed that the signal will travel through the network up to the final output layer and then return to the caller.

\begin{figure}[!h]
  \centering
  \begin{minipage}{0.65\textwidth}
    \begin{algorithm}[H]
    \caption{Forward propagation in dense layers.}\label{alg:forward_prop_dense}
    \begin{algorithmic}[1]
    \State \textbf{Input:} $I_{L_k}$ Input from one incoming connection
    \State \textbf{Output:} Output from all outgoing connected layers
    \State \textbf{Initialization:} $R  \gets Null$
    \State $N_{I} \gets $ Number of incoming connections
    \State $N_{O} \gets $ Number of outgoing connections
    \State inputs.append($I_{L_k}$)
    \If{len(inputs) < $N_{I}$}
        \State return R
    \EndIf
    \State $\bar{I} \gets \frac{1}{N_{I}} \sum_{i=0}^{N_{I}} I_{L_i}$
    \State $Z \gets Weights * \bar{I} + Bias$
    \State $A \gets ActivationFun(Z)$
    
    \For{each output connection}
        \State $L_O \gets $ Connected outgoing layer
        \State $Q_I \gets Matrix(A.rows, L_O.columns)$ 
        \State $R_{L_O} \gets L_O.ForwardPropagate((A \cdot Q_I )^{T})$ 
        \If{$R_{L_O} != Null$}
            \State $R \gets R_{L_O}$
        \EndIf
    \EndFor
    \State inputs $ \gets $ []
    \State return R
    \end{algorithmic}
    \end{algorithm}

  \end{minipage}%
  \begin{minipage}{0.35\textwidth}
  \vspace{30pt}
  \includegraphics[width=\linewidth]{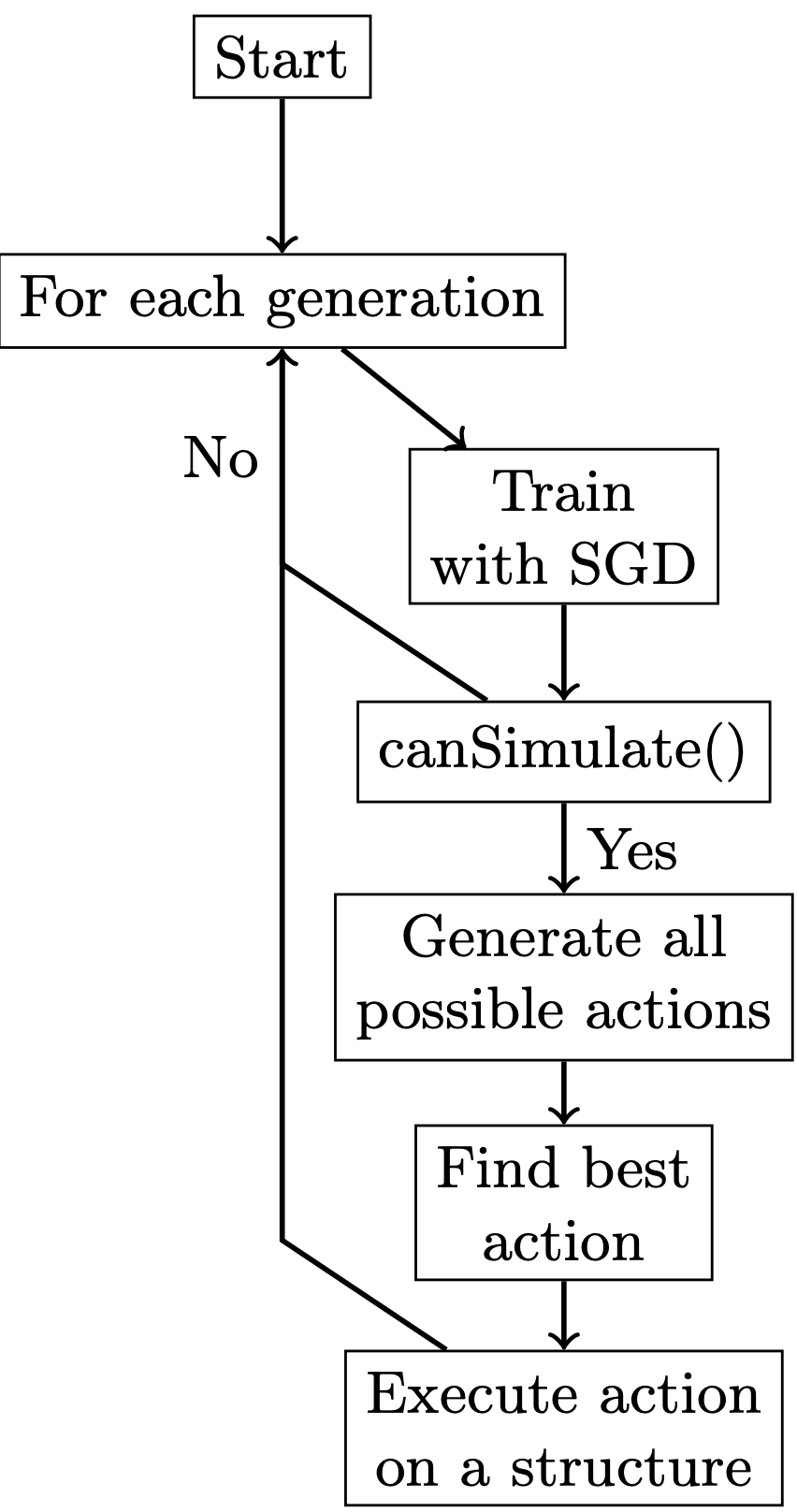}
  \end{minipage}
\end{figure}

The model starts forward propagation by sending a signal to the input layer as a forward signal. Similarly, after calculating the loss, backward propagation starts by sending a signal to the output layer as a backward signal. Forward propagation for a given layer waits until the layer receives inputs from all incoming links. The layer assumes that the received input is in the correct form. When one layer sends the input to another layer, the latter converts it to the desired size. These steps are summarized in the algorithm 2. \ref{alg:forward_prop_dense}.

After all input signals are received, the layer calculates output using learned weights and activation function. This output is adjusted to each layer on the outgoing connection by Quasi-Identity matrices ($Q_I$). A quasi-identity matrix is created to mimic identity matrices as closely as possible, by resizing the identity matrix to fit a specific size, we can quickly adjust a vector to a different size while maintaining its essential characteristics. It is an efficient way to adjust vector size without losing their fundamental features.

At this point in the algorithm, a layer can send input signals to all connected outgoing layers. This happens by running forward propagation recursively on each of the connected layers. Because the structure of the graph of connections between layers is a directional non-cyclic graph built on residual or sequential connections, we know that the last outgoing layer will return the output from the output layer of the whole model.

\subsection{Complexity Analysis}

Monte Carlo three search is time limited but it must analyze all actions at a depth of one in the searched tree. For a structure with k hidden layers, it is possible to generate $k*(k-1)/2$ actions to add sequential layers, as this is the maximum number of edges that can be added in a directed graph with k vertices without introducing cycles. The maximum number of actions for adding residual layers is $(k-1)!$, as the highest number of possible residual connections occurs in a graph resembling a path, where each vertex can connect to all subsequent vertices except itself. This means that time complexity for one generation is $O(( (k-1)! + k*(k-1)/2 ) * n * e))+C_{SGD}=O(k! * n * e) + C_{SGD}$, where $C_{SGD}$ denotes the complexity of training the model using SGD, n is the size of simulation set, e is the number of epochs for training the model in simulation. In our experiments simulation set had 10 examples per class which means that $n = 100$ and the training time in the simulation was 10 epochs.

\section{Classifying images}
\label{sec:empiricalresults}

\subsection{Datasets and empirical setup}
The experiments reported in this paper were conducted on the widely used MNIST \cite{FMNIST}  and Fashion MNIST \cite{FMNIST}  datasets. MNIST is a dataset of handwritten digits, while Fashion MNIST contains images of various items of clothing. These datasets are suitable for testing both convolutional and non-convolutional models. We have deliberately reduced the size of the initial neural architecture in order to efficiently evaluate the validity of the algorithms. The initial network configuration consists of a single convolutional input layer with a $3\times 3$ filter and a dense output layer with 10 hidden neurons. In addition, we conducted experiments using different random seeds to ensure the robustness and generalizability of our results across different training scenarios.

In what follows, we address the overall quality of the designed algorithm. Furthermore, we inspect the most critical parameters present in the method. In the conducted experiments, the training set was divided into training and testing subsets, with the testing set comprising 20\% of the training data. The distribution of images per class was even. More specifically, we used stratified sampling implemented in the $train\_test\_split$ function in the sklearn library. As a result, the testing set for MNIST comprised 8,400 images, with the training set consisting of 33,600 images. For FMNIST, the testing set included 12,000 images, and the training set comprised 48,000 images.

\subsection{Classification quality of the new approach}

The conducted experiment aimed to empirically validate two hypotheses:
\begin{description}
        \item[RH1] The first hypothesis states that the algorithm induces neural network         growth that leads to an optimal structure.
        \item[RH2] The second hypothesis states that the Monte Carlo simulation can identify         optimal changes in the network structure.
\end{description}

To verify the first hypothesis RH1, the experiments were conducted on the discussed MNIST and FMNIST datasets, with the initial network structure deliberately reduced to stimulate network growth. 

The second hypothesis RH2 was verified by comparing the Monte Carlo algorithm with greedy and random approaches. The random approach randomly selects a change to execute on the network. The greedy approach evaluates each potential change in a single step. The learning process using the Monte Carlo simulation is characterized by a stable and persistent drive to improve quality, each subsequent network change was selected to optimize the model's overall performance. After introducing a change to the network, there are a few epochs during which the network is unstable, but it quickly returns to a stable state and achieves a higher score than before the change. Because the presented data analysis problem is relatively simple and the network has a very small number of neurons, it quickly falls into a state of procrastination. In this state, the network has learned everything it could within its current structure.

\begin{figure}[htbp]
	\centering
	
	\begin{minipage}[b]{0.7\textwidth}
		\centering
		\begin{tikzpicture}
			\node[anchor=south west,inner sep=0] (image) at (0,0)
			{\includegraphics[width=\linewidth,trim={0.5cm 0cm 1.5cm 1cm},clip]{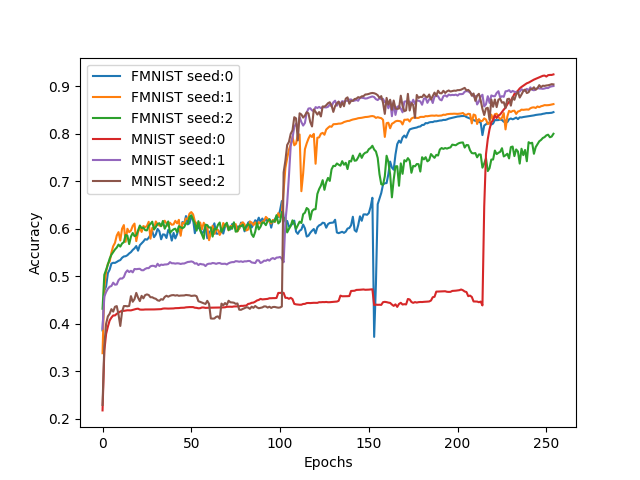}};
			\draw[->,black,thick] (2.00, 1) -- (2.00, 2.5) node[midway, right] {Stagnation};
			\draw[->,black,thick] (3.95, 1) -- (3.95, 2.5) node[midway, right, align=center] {Action\\executed};
			\draw[->,black,thick] (5.50, 1) -- (5.45, 2.0) node[midway, right, align=center] {Instability\\area};
		\end{tikzpicture}
		\caption{Neural network learning process with the Monte Carlo simulation algorithm.}
		\label{fig:exp_3_montecarlo}
	\end{minipage}%
	\hfill
	\begin{minipage}[b]{0.28\textwidth}
		\centering
		\includegraphics[width=\linewidth,keepaspectratio]{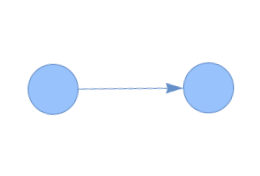} \\
		\small Epochs 0–100\\
		\includegraphics[width=\linewidth,keepaspectratio]{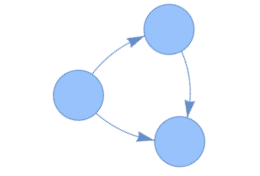} \\
		\small Epochs 100–200\\
		\includegraphics[width=\linewidth,keepaspectratio]{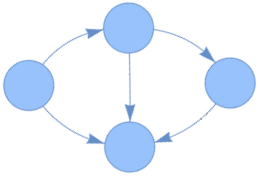} \\
		\small Epochs 200–250
		\caption{Graphs representing structure history for FMNIST seed 2.}
		\label{fig:legend}
	\end{minipage}
	
\end{figure}

In Figure \ref{fig:exp_3_montecarlo}, we observe the model learning process using our algorithm that employs Monte Carlo simulation. The graphs illustrate classification accuracy, with architectural changes in the neural network introduced every 50 epochs. Epoch number is indicated on the OX axis.

Visible ``jumps'' (see Figure \ref{fig:exp_3_montecarlo}) in the learning process occur shortly after the introduction of a modification to the network. The perturbations and instabilities during learning represent a transition phase in which the learning rate gradually changes in the first epochs after the modification. Each generation lasts for 50 epochs, so the observed phases of stability and procrastination are prominent in full cycles. The structure in each generation learns all possible features it can acquire with the structure available in the given generation. A modification was needed to extend the structure so that it could learn a new feature in the existing data set. Figure 1 shows that when the model was unable to learn more, it fell into the stability region. After the Monte Carlo simulation selects the best action, it allows the network to be better fitted. The results obtained confirm the hypothesis that the algorithm induces a growth of the neural network that leads to an optimal structure.

\begin{figure}[!ht]
  \centering
  \begin{minipage}{0.49\textwidth}
    \centering
    \includegraphics[width=\linewidth,trim={0.5cm 0cm 1cm 1cm},clip]{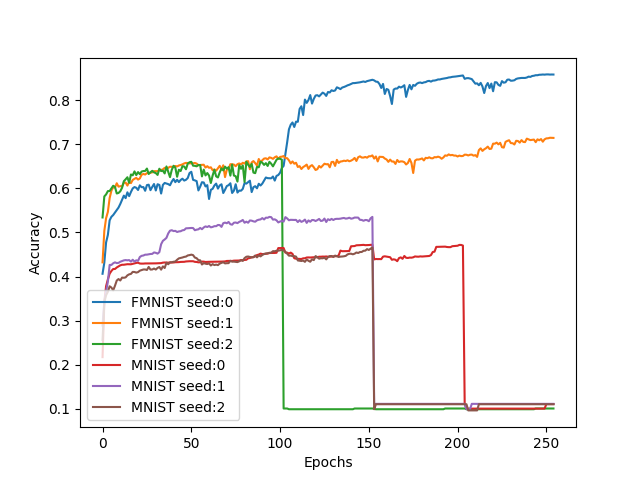}
    \caption{\centering {Ablation study: learning process with a random simulation algorithm.}}
    \label{fig:exp_random}
  \end{minipage}\hfill
  \begin{minipage}{0.49\textwidth}
    \centering
    \includegraphics[width=\linewidth,trim={0.5cm 0cm 1cm 1cm},clip]{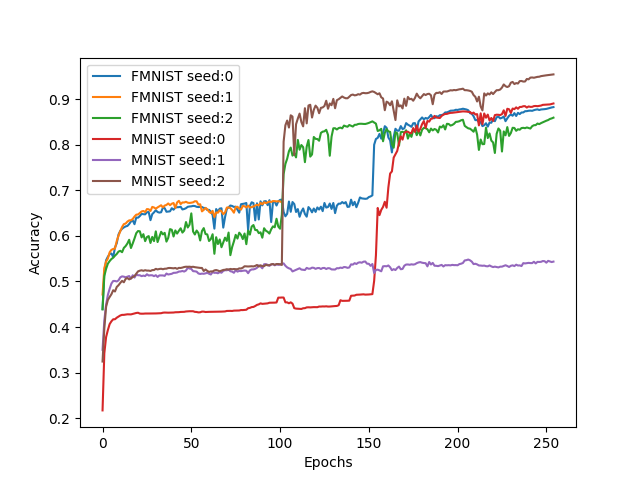}
    \caption{\centering {Ablation study: learning process with a~greedy simulation algorithm.}}
    \label{fig:exp_greedy}
  \end{minipage}
\end{figure}

Subsequently, we address the experiments conducted to verify the second research hypothesis concerning the justifiability of using the MCTS algorithm. We test two what-if scenarios. In the first scenario, we replace the MCTS method  with a random architecture modification. In the second scenario, we test a greedy search method in place of the MCTS method. The results are illustrated in Figures \ref{fig:exp_random} and \ref{fig:exp_greedy}, depicting the learning progress in terms of accuracy across epochs during training.

\begin{table}[!ht]
\centering
\caption{\centering Classification accuracy reported in a comparative study assessing the impact of a simulation algorithm. The procedure was repeated nine times with different seeds.}
\begin{tabular}{c|c|c|c|c|c|c|c|}
\cline{2-8}
& \multicolumn{3}{c|}{FMNIST}& \multicolumn{3}{c|}{MNIST} & \multirow{2}{*}{\textbf{mean}}\\
\cline{1-7}
\multicolumn{1}{|c|}{\textbf{algorithm} }& {seed 0} & {seed 1} & {seed 2} & {seed 0} & {seed 1} & {seed 2} & \\
\hline
\multicolumn{1}{|c|}{ greedy}      & 83\% & 67\% & 82\% & 86\% & 55\% & 91\% & 77.67\% \\
\multicolumn{1}{|c|}{ Monte Carlo} & 82\% & 83\% & 79\% & 90\% & 87\% & 87\% & 88.35\% \\
\multicolumn{1}{|c|}{ random }     & 83\% & 70\% & 9\% & 11\% & 11\% & 11\% & 11.43\% \\
\hline
\end{tabular}
\label{tab:meanscores}
\end{table}

In the plot concerning a randomized scenario (Figure \ref{fig:exp_random})  we observe that the learning process is highly unstable. It is possible that the algorithm may eventually reach a well-working structure. However, it is much more likely that the chosen changes will cause a total collapse of the predictive power. Observable sudden drops in the learning process indicate that the selected change prevented further learning.

In Figure \ref{fig:exp_greedy}, we illustrate the learning process for the greedy approach. It shows significantly better results than the random simulation. Subsequently, let us examine the numerical quality scores of different processing pipelines. These are summarized in Table 1. It becomes apparent that the overall score of the greedy simulation is worse than the score achieved by the Monte Carlo simulation. Changes introduced by the greedy algorithm mostly had a positive impact on the algorithm’s performance but did not lead it to achieve structures as good as those obtained through the Monte Carlo simulation.

Since the Monte Carlo simulation was able to look further into the future compared to the greedy simulation, chosen actions had a better long-term impact on the network structure. The Monte Carlo simulation selects the best change in the current generation but also considers subsequent ones, thereby determining the direction of changes in each generation leading to an optimal structure. In contrast, the greedy simulation only analyzes all possible steps in a single generation and does not consider potential future changes.

The random simulation effectively illustrates that not all changes are good for the structure, choosing the best action is crucial for the learning process, emphasizing the necessity of simulation. It is evident that random changes to the network result in the development of a structure that is unfortunately not favorable for the presented problem. This intuitively indicates that there exists an optimal structure for the given problem, and the Monte Carlo simulation is the best for discovering this structure. The obtained results confirm the hypothesis that Monte Carlo simulation can identify optimal changes in the network structure.

\subsection{Parameters of the method and their impact on the procedure}
The development of this algorithm required some design decisions that were made based on theoretical analyses and empirical results. In this section, we discuss the most important parameters that were fixed in the discussed experiments. Below, we present the results from three experiments aimed at determining key parameters for this algorithm. The first parameter determines the mode of evaluating simulation results, the second parameter dictates the initiation moment of the simulation, and the third parameter governs how the learning rate parameter should work. All experiments in this section were run on the MNIST dataset with three different seeds.

The first parameter to be tested was the mode of operation of the \textit{score function} during the algorithm simulation. The algorithm can use either the \textit{loss} or \textit{accuracy} parameter to score a model after a particular action. The choice of this parameter is particularly important when using Monte Carlo Tree Search. On the one hand, accuracy values are immediately normalized and fit well with the established and effective patterns in Monte Carlo Tree Search. On the other hand, using the loss value as an evaluation criterion has the potential to convey more information about the quality of the model after a specific action has been performed.

In that regard, we inspect and compare the impact of using loss and accuracy as two distinct evaluation criteria. The experiments were conducted on both datasets. They were repeated three times with different seeds.

\begin{table}[!ht]
\centering
\caption{Classification accuracy depending on various score function modes. }\label{tab:scorefmode}
\begin{tabular}{|c||c|c|c||c|}
\hline
\textbf{score function} &\textbf{seed 1} &\textbf{seed 2} &\textbf{seed 3} &\textbf{mean}\\
\hline
accuracy & 89.5\% & 85.0\% & 53.7\% & 76.0\% \\
loss & 44.9\% & 89.5\% & 55.7\% & 63.3\% \\
\hline
\end{tabular}
\end{table}

The experiments have shown that the mean score of the models that used accuracy in grading actions was bigger than that of those that used loss, which is shown in Table  \ref{tab:scorefmode}. We stipulate that it is because UCB1 (UCB stands for Upper Confidence Bounds) in Monte Carlo simulation works better with normalized values like accuracy.

Subsequently, we examined different operational modes for the simulation orchestrator. There are several strategies one can design for this task. We chose to investigate three specific modes: \textit{constant}, \textit{progress check}, and \textit{overfit}.\vspace{-5pt}

\begin{table}[!ht]
\centering
\caption{\centering Classification accuracy and the number of changes in network architecture for various simulation orchestrator working modes. ``sd'' stands for seed.}\label{tab:scheduler}
\begin{tabular}{|c||c|c|c||c|c|c|c| |c|}
\hline
& \multicolumn{4}{c|}{accuracy} & \multicolumn{4}{c|}{ number of simulations} \\
\cline{1-9}
\textbf{orchestrator}  &\textbf{sd1} &\textbf{sd2} &\textbf{sd3} &\textbf{mean} &\textbf{sd1} &\textbf{sd2} &\textbf{sd3} &\textbf{mean}\\
\hline
overfit         & 89.3\% & 86.6\% & 84.6\% & 86.3\% & 18 & 18 & 18 & 18 \\ 
progress check  & 83.6\% & 88.0\% & 90.6\% & 87.4\% & 10 & 7  & 9 & 8.6 \\
constant        & 89.9\% & 88.2\% & 87.1\% & 88.3\% & 18 & 18 & 18 & 18 \\
\hline
\end{tabular}
\end{table}

Table \ref{tab:scheduler} showcases the average classification accuracy on test sets for various orchestrator operation modes. The ``overfit'' method displayed slightly worse performance compared to the other two. In this mode, the algorithm triggers simulations when there is a high probability of overfitting based on the model's learning history. The ``constant'' mode conducts a simulation in each generation, while the ``progress check'' mode verifies if the model has achieved improved accuracy compared to the previous generation; if not, a simulation is initiated. All these methods show promise, yielding similar results. Despite the ``constant'' method having the best mean score, the “progress check” mode attains the highest maximum score. Consequently, we analyzed the extent of changes induced by these orchestrators. The primary objective of this module is to find a balance between exploiting and exploring the model structure and adjusting the number of changes to the model’s learning progress. Table \ref{tab:scheduler} details the number of changes caused by the orchestrator. Both the ``overfit'' and “constant'' orchestrators made changes in each generation. Perhaps the ``overfit'' mode should be better parameterized for a given problem to be less or more sensitive. The ``progress check'' made changes in approximately half of the possible generations. We reduced the number of neurons in the layers to close to 10, explaining why adding more layers consistently yielded high scores in all orchestrators. However, the ``progress check'' orchestrator, with fewer changes, emerged as the most stable and efficient method, making it the likely best choice for future experiments.

\begin{table}[!ht]
\centering
\caption{\centering Classification accuracy depending on different  learning rate scheduler modes.}\label{tab:lrscheduler}
\begin{tabular}{|c||c|c|c||c|}
\cline{1-5}
\textbf{Learning rate scheduler} &\textbf{seed 1} &\textbf{seed 2} &\textbf{seed 3} &\textbf{mean}\\
\hline
constant mode & 54.3\% & 83.9\% & 81.9\% & 73.3\% \\
progress mode & 86.5\% & 87.9\% & 85.2\% & 86.5\% \\
\hline
\end{tabular}
\end{table}

Subsequently, we compared the use of a progressive learning rate with a constant learning rate. Progressive means that the learning rate increases up to a certain point during one generation and then decreases almost to zero before starting the next generation. This happens so that the changes introduced by the algorithm to the model have the least impact on the features learned so far.  With a constant learning rate, the same value applies throughout all generations.  From the conducted experiments, we inferred that the progressive mode yields higher results.

\section{Classifying multivariate time series}
\label{sec:mtsc}
The developed method shows significant potential for addressing the problem of multivariate time series classification \cite{JOCS_TIMESERIES}. A specialized initial structure has been designed specifically for time series problems.

A single hidden layer was added between the input and output layers. It has a vital role because if it were not there, the input signals would always merge with other signals only in the output layer, no matter how much the whole network would expand. At this point, actions allow expand the structure for each series separately and together. Each series will expand separately between the input layer and the hidden layer.

This unique structure makes it possible to independently develop neural networks for each time series, allowing full use of information from each variable separately. With this approach, the method can effectively deal with the diverse characteristics of each time series, providing comprehensive analysis and interpretation of classification results.

\subsection{Data transformations}

Datasets involved in the experiments presented in this paper were thoroughly addressed in the paper ``The Great Multivariate Time Series Classification Bake Off'' \cite{Ruiz2021}.

In the first step, we processed each series to obtain \textit{recurrence plots}. It is a prime time series imaging method \cite{RecrencepLots}.

\begin{figure}
    \begin{subfigure}[b]{0.5\textwidth}
        \includegraphics[width=\linewidth]{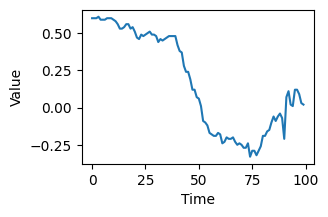}
        \caption{\centering  Variable \#1, time series \#1, Epilepsy dataset.}
        \label{fig:ts_0_0}
    \end{subfigure}
    \begin{subfigure}[b]{0.5\textwidth}
        \centering
        \includegraphics[width=0.7\linewidth]{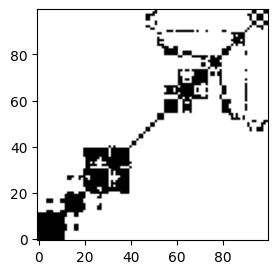}
        \caption{\centering Recurrence plot for variable \#1, time series \#1.}
        \label{fig:ts_0_1}
    \end{subfigure}
    \newline
    
    \begin{subfigure}[b]{0.5\textwidth}
        \includegraphics[width=\linewidth]{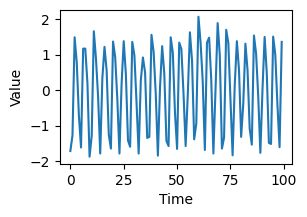}
        \caption{\centering Variable \#2, time series \#1, Epilepsy dataset.}
        \label{fig:ts_1_0}
    \end{subfigure}
    \begin{subfigure}[b]{0.5\textwidth}
        \centering
        \includegraphics[width=0.7\linewidth]{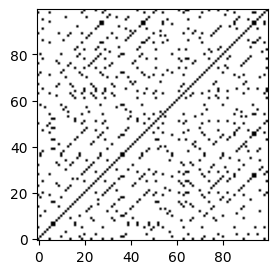}
        \caption{\centering Recurrence plot for variable \#2, time series \#1.}
        \label{fig:ts_1_1}
    \end{subfigure}

    \begin{subfigure}[b]{0.5\textwidth}
        \includegraphics[width=\linewidth]{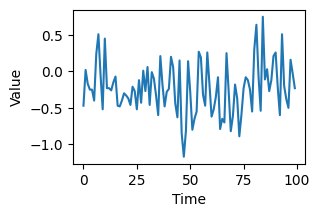}
        \caption{\centering  Variable \#3, time series \#1, Epilepsy dataset.}
        \label{fig:ts_2_0}
    \end{subfigure}
    \begin{subfigure}[b]{0.5\textwidth}
        \centering
        \includegraphics[width=0.7\linewidth]{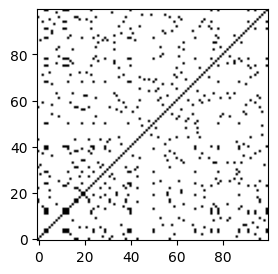}
        \caption{\centering  Recurrence plot for variable \#3, time series \#1.}
        \label{fig:ts_2_1}
    \end{subfigure}
    
    \caption{\centering  {   First time series from the multivariate time series Epilepsy dataset, with   recurrence plots generated from each time series. Each time series consists of   three variables. This time series belongs to the class named ``Epilepsy''. }}
    \label{fig:timeseries}
\end{figure}

To demonstrate how recurrence plots work, we present two examples (two time series) from the Epilepsy dataset. Each time series in this set is composed of three variables. The original three variables of time series \#1 are depicted with the usual time-value plots in Figures \ref{fig:ts_0_0}, \ref{fig:ts_1_0}, and \ref{fig:ts_2_0}). The same three variables were transformed to recurrence plots and depicted in Figures ,  , and .
 Analogous plots were prepared for time series \#2. They are visible in Figures , , and  -- time series value in time. Figures , , and  show these three variables in the form of recurrence plots.

The recurrence plot-based representation helps visualize the data dynamic structure and patterns. This representation model matches well with convolutional layers, which are, at the same time, the designed input in our proposed dynamic neural model. The formula for recurrence plots (see Eqn. (\ref{eqn:recurrenceplot})) checks how close two points in a system are by measuring the distance between them. If the distance between states is less than or equal to epsilon, it assigns a 1. Otherwise, it assigns a 0. This creates a matrix that shows when similar states happen over time.

\begin{equation}\label{eqn:recurrenceplot}
	R_{i,j} = 
	\begin{cases}
		1 & \quad \textrm{if } \| \mathbf{x}_i - \mathbf{x}_j \| \leq \varepsilon \\
		0                 & \quad \textrm{otherwise}              
	\end{cases}
\end{equation}

Figure \ref{fig:timeseries} illustrates the first time series from the multivariate Epilepsy dataset\footnote{The data can be downloaded from: \\ \url{https://www.timeseriesclassification.com/description.php?Dataset=Epilepsy}}. This time series belongs to the class named ``Epilepsy'', which corresponds to a recording of a seizure. Each time series is composed of three variables. In Figure \ref{fig:timeseries}, we show three variables from the first time series plotted in two settings. On the left-hand side, we show plain value in time plot. On the right-hand side, we show recurrence plots.

The method created worked only on the CPU, which induced performance problems. Therefore, it was not possible to analyze much larger datasets, less demanding sets were chosen. This, unfortunately, has a big impact on the results, because in many cases there was a certain polarization: with simpler data the algorithm reached great results, close to 100\%, with more difficult data it stopped at an average of 60\%. Details of the datasets used in the experiment are presented in Table \ref{tab:datasetlstmparameters}.  For each dataset, the table provides several key attributes. The Size of the dataset indicates the number of samples in both the training and testing sets. Dimensions refer to the number of features or variables describing each data point. Length represents the time series length, indicating the number of time

\begin{table}[h]
\centering
\caption{Parameters of the multivariate time series used. ``Dm'' stands for Dimensions, ``Ln'' stands for Length, ``Cl'' stands for the number of classes.} \label{tab:datasetlstmparameters}
\begin{tabular}{|c|c|c|c|c|c|}
\hline
\multirow{2}{*}{\textbf{Name}} & \multicolumn{2}{|c|}{\textbf{Size of the dataset}} & \multirow{2}{*}{\textbf{Dm} }& \multirow{2}{*}{\textbf{Ln} }&\multirow{2}{*}{\textbf{ Cl }}\\
\cline{2-3}
  &  \textbf{Train} &   \textbf{Test} &   &   &   \\
\hline
ArticularyWordRecognition & 275        & 300       & 9    & 144      & 25      \\
AtrialFibrillation        & 15         & 15        & 2    & 640      & 3       \\
BasicMotions              & 40         & 40        & 6    & 100      & 4       \\
Epilepsy                  & 137        & 138       & 3    & 206      & 4       \\
ERing                     & 30         & 270       & 4    & 65       & 6       \\
Libras                    & 180        & 180       & 2    & 45       & 15      \\
LSST                      & 2459       & 2466      & 6    & 36       & 14      \\
NATOPS                    & 180        & 180       & 24   & 51       & 6       \\
RacketSports              & 151        & 152       & 6    & 30       & 4       \\
FingerMovements           & 316        & 100       & 28   & 50       & 2       \\
StandWalkJump             & 12         & 15        & 4    & 2500     & 3       \\
UWaveGestureLibrary       & 120        & 320       & 3    & 315      & 8       \\
\hline
\end{tabular}
\end{table}

In our method, the input layers are made up of convolutional layers that process recurrence plots. Our models architecture resembles InceptionTime \cite{IsmailFawaz2020} because it also starts with convolutional layers and includes residual connections. These connections help the network skip over certain layers, which prevents performance issues in deeper networks and ensures efficient learning. We believe our algorithm can be especially useful in tasks similar to those handled by InceptionTime.

\subsection{Empirical analysis}
The primary objective of this experiment was to assess the potential of developing neural networks that process each time series independently. This experiment aims to compare the performance of a neural network with a single input layer to that of a neural network with multiple input layers, with both architectures optimized using a newly developed algorithm. In the single input layer approach, each time series is transformed into a recurrence plot and fed into the convolutional input layer as separate channels. The multi-input layer network assigns data from each individual time series to distinct input layers, allowing each layer to evolve independently and adapt specifically to its corresponding time series.

\begin{figure*}[t!]
    \centering
    \begin{subfigure}[t]{0.5\textwidth}
        \centering
\includegraphics[width=\linewidth]{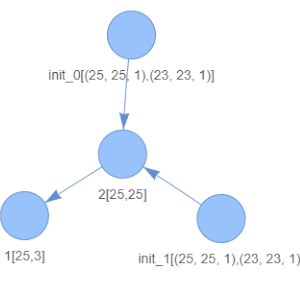}
        \caption{\centering {Architecture in the \newline first generation}}
    \end{subfigure}%
    ~ 
    \begin{subfigure}[t]{0.5\textwidth}
        \centering
       \includegraphics[width=\linewidth]{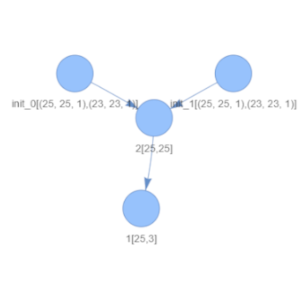}
        \caption{\centering {Architecture in the \newline third generation}}
    \end{subfigure}\\
    
    \begin{subfigure}[t]{0.5\textwidth}
        \centering
\includegraphics[width=\linewidth]{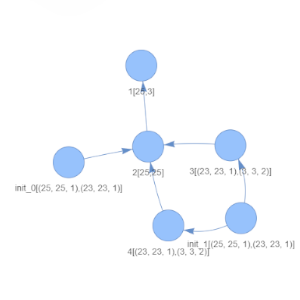}
\caption{\centering {Architecture in the \newline sixth generation}}
    \end{subfigure}%
    ~ 
    \begin{subfigure}[t]{0.5\textwidth}
        \centering
\includegraphics[width=\linewidth]{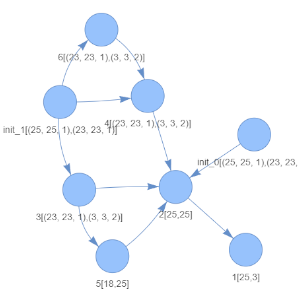}
\caption{{Architecture in the \newline tenth generation}}
    \end{subfigure}    
    \caption{ \centering 
    {Structure development history for the multivariable time series     	problem on the AtrialFibrillation dataset seed=3 for multiple input layers}}
    \label{img:manyin_false}
\end{figure*}

\begin{figure*}[t!]
    \centering
    \begin{subfigure}[t]{0.5\textwidth}
        \centering
        \includegraphics[width=\linewidth]{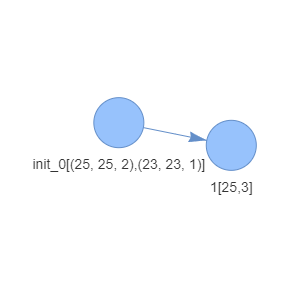}
        \caption{\centering Architecture in the \newline first generation}
    \end{subfigure}%
    ~ 
    \begin{subfigure}[t]{0.5\textwidth}
        \centering
        \includegraphics[width=\linewidth]{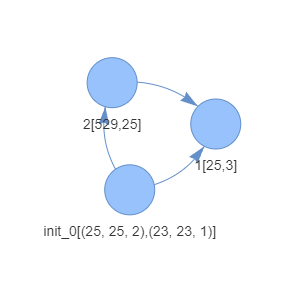}
        \caption{\centering Architecture in the \newline third generation}
    \end{subfigure}\\

    \begin{subfigure}[t]{0.5\textwidth}
        \centering
        \includegraphics[width=\linewidth]{manyin_true_3.png}
        \caption{\centering Architecture in the \newline sixth generation}
    \end{subfigure}%
    ~ 
    \begin{subfigure}[t]{0.5\textwidth}
        \centering
       \includegraphics[width=\linewidth]{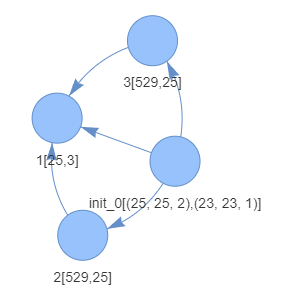}
        \caption{\centering Architecture in the \newline tenth generation}
    \end{subfigure}
\caption{\centering {Structure development history for a multivariable time series problem 		on an AtrialFibrillation dataset seed=3 for a single input layer}}\label{img:manyin_true}
\end{figure*}

The presented figures in \ref{img:manyin_false}, \ref{img:manyin_true}  illustrate the evolutionary history of structural modifications made to the neural network architecture. The networks were trained on the AtrialFibrillation dataset, undergoing a training process consisting of 10 generations, each with 50 epochs. After each generation, the network structure was allowed to evolve. The figures specifically depict the network architectures at the first, third, sixth, and tenth generations, highlighting the progression of changes over the course of training.

\begin{table}[h]
\centering
\caption{\centering Classification accuracy on the training set, depending on whether a 	separate input layer is used for each series or a single input layer is applied to 	all series.}\label{tab:exp4_3}
\begin{tabular}{|c|c|c|c|}
\hline
\multirow{2}{*}{\textbf{Dataset}} & \multirow{2}{*}{\textbf{Seed}} & \multicolumn{2}{|c|}{\textbf{Type of input layers}} \\
\cline{3-4}
 &   & \textbf{Single Layer}& \textbf{Layer per dim} \\
\hline

\multirow{2}{*}{ArticularyWordRecognition} 
  & 0 & 32.46\% & 72.86\% \\
  & 1 & 49.95\% & 69.52\% \\
\hline
\multirow{2}{*}{AtrialFibrillation} 
  & 0 & 82.16\% & 78.11\% \\
  & 1 & 58.33\% & 78.92\% \\
\hline
\multirow{2}{*}{BasicMotions} 
  & 0 & 100.0\% & 100.0\% \\
  & 1 & 100.0\% & 100.0\% \\
\hline
\multirow{2}{*}{ERing} 
  & 0 & 99.83\% & 91.08\% \\
  & 1 & 95.42\% & 92.75\% \\
\hline
\multirow{2}{*}{Epilepsy} 
  & 0 & 100.0\% & 99.32\% \\
  & 1 & 100.0\% & 99.32\% \\
\hline
\multirow{2}{*}{Libras}
  & 0 & 97.50\% & 97.19\% \\
  & 1 & 96.97\% & 94.97\% \\
\hline
\multirow{2}{*}{LSST } 
  & 0 & 40.52\% & 49.26\% \\
  & 1 & 40.52\% & 49.26\% \\
\hline
\multirow{2}{*}{NATOPS} 
  & 0 & 92.71\% & 98.16\% \\
  & 1 & 96.92\% & 79.17\% \\
\hline
\multirow{2}{*}{RacketSports} 
  & 0 & 100.0\% & 98.01\% \\
  & 1 & 100.0\% & 97.48\% \\
\hline

\hline
\textbf{Mean} & -- & 82.41\% & 85.88\% \\
\hline
\end{tabular}

\end{table}

Figures \ref{img:manyin_false} illustrate the network’s initial configuration, consisting of two input layers and a single hidden layer. This distinctive architecture enables the network to evolve in subsequent stages, allowing for the independent analysis of individual time series. In contrast, Figures 15 depict a configuration with a single input layer, which is a convolutional layer designed to process multiple time series concurrently, similar to the way RGB channels are handled in image data.

The structure utilizing multiple input layers ( Figure \ref{img:manyin_true}) evolved quite differently compared to the single input layer architecture ( Figure \ref{img:manyin_false}). It can be observed that most modifications occurred in the layer corresponding to time series with index 1. In this example, one sequential layer and three residual layers were added, all dedicated to analyzing only series 1. This occurred because changes made to the part dealing with series 1 had the most significant impact on improving the network's performance. These changes reduced the need for further modifications, and as the network reached a state of stagnation, simulations indicated that alterations in the series 1 section were most beneficial for the existing structure. The resulting architecture can be compared to the single input layer model shown in Figure \ref{img:manyin_false}. It developed gradually through incremental changes and closely resembles the structure found when analyzing the MNIST dataset. Since this structure has consistently appeared as an optimal configuration in previous experiments, it requires further testing to confirm its effectiveness and applicability.

Table \ref{tab:exp4_3} shows the training set accuracy when comparing architectures that have one layer per time series with those that use a single input layer. The multilayer approach generally provides better average performance. A key feature of the algorithm is its ability to handle multivariate time series by using a separate network architecture for each time series. Additionally, the overall effectiveness of this algorithm is strong compared to widely used methods.

In the presented algorithm if a certain time series represents a variable that is crucial for prediction then the network branch associated with that series will expand more than the others. This hypothesis was validated in empirical
experiments.

Table \ref{tab:wyniki_wybrane} presents a comparison of our algorithm’s results with those of various established methods for multivariate time series classification, as discussed in~\cite{Ruiz2021}.  The performance of the proposed method on the test dataset was compared to the average results of four popular techniques. Our method was evaluated against the following methods: Dynamic Time Warping (DTW) \cite{JOCS_DTW},  Random Convolutional Kernel Transform (ROCKET), Canonical Interval Forest (CIF) \cite{MCTS_CIF}, and HIVE-COTE \cite{MCTS_HIVECOTE}.

\begin{table}[h]
\centering
\caption{Classification accuracy on the test set, comparing the best results of
	our method with other popular methods on selected datasets.} \label{tab:wyniki_wybrane}
\begin{tabular}{|c|c|c|c|c|c|}
\hline
& {DTWD} & {ROCKET}  & {CIF} & {HIVE-COTE} & {GROWINGNN} \\ \hline
AtrialFibrillation & 23.56\% & 24.89\% & 25.11\% & 29.33\% &  \textbf{34.18\%}\\
BasicMotions & 95.25\% & 99.00\% & 99.75\% & \textbf{100.0\%} & 99.73\%\\
Epilepsy & 96.30\% & 99.08\% & 98.38\% & \textbf{100.0\%} & 98.31\%\\
ERing & 92.91\% & \textbf{98.05\%} & 95.65\% & 94.26\% & 26.24\%\\
FingerMovements & 54.17\% & 55.27\% & 53.90\% & 53.77\% & \textbf{57.40\%}\\
Libras & 88.04\% & 90.61\% & \textbf{91.67\%} & 90.28\% & 79.36\%\\
LSST & 54.76\% & \textbf{63.15\%} & 56.17\% & 53.84\% & 37.51\%\\
NATOPS & 82.04\% & \textbf{88.54\%} & 84.41\% & 82.85\% & 43.93\%\\
RacketSports & 85.64\% & 92.79\% & 89.30\% & \textbf{90.64\%} & 78.55\%\\
StandWalkJump & 22.00\% & 45.56\% & 45.11\% & 40.67\% & \textbf{49.19\%}\\
UWaveGestureLibrary & 92.28\% & \textbf{94.43\%} & 92.42\% & 91.31\% & 35.26\%\\
\hline
\end{tabular}
\end{table}

\section{Conclusion}
\label{sec:conclusion}

The paper has brought forward a new approach to dynamic neural network training procedures. The proposed procedure relies on a simulation orchestrator that launches an MCTS procedure. The outcome of this procedure is a decision concerning a change in the neural architecture. The addressed solution works on the level of a layer: after each simulation, we may decide to add a layer, remove a layer, or keep the current architecture intact. The detailed formalism was presented for convolutional, plain sequential (dense) layers, and residual sequential layers. 
The new method, in contrast to the approaches existing in the literature, empowers more flexible model design through the use of a wide variety of neural layers.

The validity of the use of the MCTS algorithm for design was tested in an ablation and substitution study. We have replaced the Monte Carlo simulations with a random decision-making algorithm and with a greedy algorithm. The latter performed substantially worse. The random algorithm was unacceptable.

The proposed method shows strong potential for multivariate time series classification, outperforming several popular algorithms on test sets from three different datasets. The method is particularly effective when employing multiple input layers, algorithm is capable of designing neural network structures customized for each individual time series, improving overall performance. While the algorithm demonstrates impressive results, further optimization is required to enhance its efficiency on more complex datasets. Overall, the algorithm is promising and warrants additional research to fully explore its capabilities.

An indispensable component of the delivered study was the prepared source code. It is openly available as a Python package named $growingnn$. It was uploaded to PyPi: \url{https://pypi.org/project/growingnn/}. It contains the implementation of the training algorithm prepared from scratch in Python. We want to underline the scarcity of open-source codes in the domain of dynamic neural topology adjustment methods and we believe that our work would bring practical value to the researchers working in this area.


\end{document}